%% file: main.tex
\documentclass{article} %
\usepackage{iclr2027_conference,times}
\usepackage[most]{tcolorbox}

\input{math_commands.tex}

\usepackage{hyperref}
\usepackage{url}

\input{preamble}

\title{Rethinking Probability-Based Reinforcement Learning From Posterior Concentration}

\iclrfinalcopy
\author{%
\begin{minipage}[t]{0.96\textwidth}
\centering
\textbf{
Shiu-Hong Kao$^1$,
Yubo Zhao$^2$,
Zhenyu Tian$^3$,
Pengzhan Sun$^1$,
Yicong Li$^{3,\star}$,
Angela Yao$^1$
}
\\[0.4em]
\normalfont\small
$^1$National University of Singapore
\\
$^2$The Hong Kong University of Science and Technology
\\
$^3$University of Science and Technology of China
\\
$^\star$Corresponding author
\end{minipage}
}

\usepackage{soul}

\usepackage{lipsum}

\begin{document}

\maketitle
\begin{abstract}
Verifier-free reinforcement learning with probability-based rewards offers a promising way to train LLMs on general reasoning tasks where external verifiers are unavailable. Yet the reliability of these rewards, especially in long-horizon reasoning, remains underexplored. This work identifies a length-dependent failure mode of probability rewards, which we call the Posterior Concentration Phenomenon (PCP). We show that 
the probability of a reference answer conditioned on a reasoning trace 
{often collapses to a low-variance interval}
as the trace becomes lengthy. 
This phenomenon results in nearly indistinguishable rewards, which, under GRPO-based settings, %
makes 
probability-based policy optimization
unstable and %
inefficient. 
Motivated by this, we propose \textit{Reinforcement Learning with Concentration-aware Posterior Rewards} \textbf{(RLCPR)}, a verifier-free RL framework to explicitly account for PCP for better optimization stability and token efficiency.
It has two %
components: \textbf{uncertainty-aware data sampling}, which reduces concentration-prone rollouts before generation, and \textbf{concentration-aware regularization}, which penalizes unnecessarily long traces when posterior rewards collapse. {Extensive experiments show that, alongside higher token efficiency, RLCPR outperforms the state-of-the-art verifier-free RL baseline by up to 4.0\% on six of seven benchmarks, including general-domain and mathematical reasoning challenges.}

\end{abstract}
\vspace{-0.05in}
\section{Introduction}\label{sec:introduction}
\input{1_introduction}
\section{Related Work}
\input{2_related_work}

\section{Preliminaries}

\input{2.5_preliminary}

\section{Posterior Concentration Phenomenon}~\label{sec:3_theory}
\input{3_theory}

\vspace{-0.5em}
\section{RLCPR}\label{sec:4_method}
\vspace{-0.5em}
\input{4_method}
\vspace{-0.5em}
\section{Experiments}\label{sec:5_experiments}
\input{5_experiments}

\vspace{-0.5em}
\section{Conclusion}\label{sec:6_conclusion}
\vspace{-0.5em}
\input{6_conclusion}

\section*{AI Use Statement}
We used AI tools to polish the manuscript’s language and presentation. The theoretical ideas and concepts were developed by the authors, with AI assistance used to check proofs for potential errors and refine their exposition. AI agents also assisted with code implementation and figure creation. The authors retain full responsibility for the correctness, originality, and integrity of the work.

\bibliography{iclr2026_conference}
\bibliographystyle{iclr2027_conference}

\appendix
\input{appendix}

\end{document}

%% file: math_commands.tex
\usepackage{amsmath,amsfonts,bm}

\def\eqref#1{equation~\ref{#1}}

\def\1{\bm{1}}

\DeclareMathAlphabet{\mathsfit}{\encodingdefault}{\sfdefault}{m}{sl}
\SetMathAlphabet{\mathsfit}{bold}{\encodingdefault}{\sfdefault}{bx}{n}

\DeclareMathOperator*{\argmax}{arg\,max}

%% file: preamble.tex
\usepackage{bbm}
\newcommand{\Cond}{\,|\,}

\newtheorem{lemma}{Lemma}[section]
\newtheorem{proposition}{Proposition}[section]
\usepackage{cleveref}
\usepackage{graphicx}
\Crefname{equation}{Eq.}{Eqs.}
\Crefname{theorem}{Thm.}{Thms.}
\Crefname{lemma}{Lem.}{Lems.}
\Crefname{proposition}{Prop.}{Props.}
\Crefname{section}{Sec.}{Secs.}
\Crefname{figure}{Fig.}{Figs.}
\usepackage{wrapfig}
\usepackage{subcaption}
\usepackage{mathtools}
\usepackage{multicol}
\usepackage{multirow}
\usepackage{booktabs}
\usepackage[table]{xcolor}
\usepackage[dvipsnames]{xcolor}
\usepackage{amssymb}%
\usepackage{pifont}%
\usepackage{arydshln} 
\newcommand{\thicktoprule}{\specialrule{1pt}{0pt}{0pt}}
\newcommand{\thickbottomrule}{\specialrule{1pt}{0pt}{0pt}}
\newcommand{\gain}[1]{\,{\scriptsize\textcolor{green!45!black}{(+#1)}}}

\newtcblisting{promptbox}[1]{%
  enhanced,
  listing only,
  colback=white,
  colframe=black!75,
  colbacktitle=black!75,
  coltitle=white,
  fonttitle=\bfseries\sffamily\small,
  title={#1},
  boxrule=1.3pt,
  arc=5pt, outer arc=5pt,
  left=6pt, right=6pt, top=3pt, bottom=3pt,
  toptitle=2pt, bottomtitle=2pt,
  listing options={
    basicstyle=\ttfamily\scriptsize,
    backgroundcolor=\color{white},
    breaklines=true,
    breakatwhitespace=true,
    showstringspaces=false,
    columns=fullflexible,
    keepspaces=true
  }
}

%% file: 1_introduction.tex
\begin{wrapfigure}{rt}{0.6\linewidth}
    \centering
    \includegraphics[width=\linewidth]{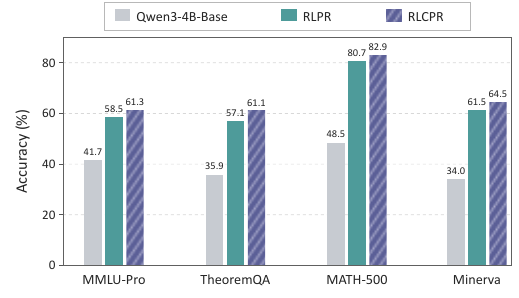}
    \caption{RLCPR performance on representative general-domain and mathematical benchmarks.}
    \label{fig:teaser2}
\end{wrapfigure}

Reinforcement learning (RL) has become a prominent solution to advance the reasoning capabilities of large language models (LLMs)~\citep{RLHF}. By optimizing models on sampled reasoning trajectories, RL lets a model learn from its own successes and failures, and has delivered strong gains in mathematical reasoning~\citep{hu2026open,zeng2025simplerl,yu2026dapo}, code generation~\citep{cui2025process}, and complex decision-making tasks. %

However, many existing RL approaches for LLM reasoning rely on verifiable rewards, where correctness can be easily checked by pre-defined rules or by external verifier-purposed models~\citep{lambert2024tulu,chen2026does}. This requirement limits their applicability to broader reasoning settings where ground-truth verification is unavailable, ambiguous, or expensive. Recently, probability-based rewards have emerged as a promising verifier-free alternative~\citep{yu2025rlpr,liu2025nover,zhou2025reinforcing}. Instead of using an external verifier, these methods directly use the LLM’s own probability estimates to construct rewards. A representative formulation is the posterior reward, which scores a reasoning trace ($z$) by the model’s posterior probability of generating the ground-truth answer ($y^*$) after conditioning on $z$, i.e., $R_\theta(z)=P(y^*\Cond x,z;\theta)$. Intuitively, this posterior probability-based reward encourages the model to learn %
reasoning traces %
associated with a higher %
probability %
of %
the reference answer. %
It provides dense, self-contained supervision that extends beyond strictly verifiable domains and requires no external model during training.

Despite their practical appeal, the mechanism behind posterior probability-based rewards is underexplored.
It is unclear whether these rewards can reliably distinguish high- versus low-quality reasoning traces. This question is particularly important for group-relative RL algorithms such as GRPO~\citep{grpo}, where training depends on the difference in the rewards across multiple rollouts for the same prompt.
Recent evidence suggests a reason for concern. \cite{welch2026cost} show that, once a model has produced a reasoning trace, the likelihood it assigns to an answer increasingly reflects consistency with that trace rather than the model's belief in the answer itself. 
If posterior rewards inherit this behavior, a sufficiently long trace may lock in its answer regardless of whether the reasoning is sound.

Building on this concern, this work identifies a length-dependent failure mode of posterior rewards, which we call the \emph{Posterior Concentration Phenomenon} (PCP). Our exploration experiments show that, as reasoning traces grow longer, posterior rewards for traces ending in correct and incorrect answers become less distinguishable, while their cross-rollout variance also contracts. We further connect this empirical behavior to recent results on log-likelihood concentration in long LLM generations \citep{bell2024slaves,yang2026llmprobabilityconcentrationalignment}. Specifically, 
as sequence likelihoods concentrate, it causes the posterior probability of the reference answer to depend less on the specific reasoning trajectory. Under a mild condition motivated by empirical evidence on LLM branching factors~\citep{yang2026llmprobabilityconcentrationalignment}, we show that different reasoning traces become increasingly likely to receive similar posterior rewards as the considered range of reasoning lengths expands.
This reveals a fundamental limitation of posterior rewards for group-relative optimization: as their ability to reflect reasoning quality diminishes, the reward differences that GRPO relies on no longer indicate which rollouts are better, leaving the policy without an informative optimization direction.
Worse still, since rollouts tend to grow longer as training proceeds, the training dynamics are pushed deeper into the concentrated regime over time. 
PCP reframes how posterior-based RL should be designed. If posterior rewards lose discriminative power as traces lengthen, a training framework should (i) avoid spending generation budget on prompts likely to produce concentrated rollout groups, and (ii) prevent the policy from drifting into the long-trace regime where posterior rewards no longer reflect reasoning quality. Existing methods address neither directly. RLPR~\citep{yu2025rlpr}, for instance, discards rollout groups with low reward variance, but it does so without characterizing when or why such groups arise. Moreover, because filtering occurs only after rollouts are generated and scored, the generation cost of the discarded groups is wasted rather than avoided.
Motivated by our analysis, we propose \textbf{R}einforcement \textbf{L}earning with \textbf{C}oncentration-aware \textbf{P}osterior \textbf{R}ewards (\textbf{RLCPR}), {a verifier-free RL framework that explicitly accounts for PCP for better optimization stability and token efficiency.} RLCPR contains two major components:
(1) an uncertainty-aware data sampling method to efficiently sample informative training data with a lower risk of PCP, (2) a regularization mechanism that explicitly accounts for PCP and penalizes unnecessarily lengthy reasoning traces. Rather than replacing the posterior rewards, RLCPR improves their reliability by preserving useful posterior information while reducing the adverse effect of long-horizon concentration. {Overall, RLCPR outperforms RLPR by up to 4.0 percentage points on six of seven benchmarks, including MMLU-Pro, TheoremQA, WebInstruct, MATH-500, Minerva, and AIME24.}
Our contributions are summarized as follows:
\begin{itemize}
    \item We identify the Posterior Concentration Phenomenon (PCP), a length-dependent loss of answer-probability discriminability in long LLM reasoning traces, and connect it to the reward collapse in probability-based RL.

\item We provide the first theoretical formulation of posterior-reward collapse and support it empirically through diminishing reward separation and contracting reward variance.

\item We propose RLCPR, which explicitly accounts for PCP through uncertainty-aware data sampling and concentration-aware regularization.
\end{itemize}

%% file: 2_related_work.tex
\noindent{\bf Reasoning and confidence reliability.}
Prior work examines whether language models' confidence reflects answer
correctness~\citep{jiang2021can,kadavath2022language}. Explicit reasoning
complicates this relationship. While \citet{yoon2026reasoning} find that
verbalized confidence can become better calibrated as chain of thoughts (CoT) unfolds,
\citet{welch2026cost} show that answer-token probabilities in
vision-language models can increasingly reflect consistency with the
generated reasoning rather than correctness. This concern is reinforced
by studies of reasoning faithfulness: CoT explanations can rationalize
hidden biases~\citep{turpin2023language}, and final predictions do not
always depend on the stated reasoning~\citep{lanham2023measuring}.
Moreover, prior beliefs can induce confirmation bias in both rationale
generation and answer prediction~\citep{wan2025unveiling}. These findings
motivate examining whether confidence reliably distinguishes reasoning
quality. Our work studies this question in posterior-based RL, where
rewards must remain \emph{relatively discriminative} across reasoning
traces, and identifies a length-dependent loss of this property through PCP.

\noindent{\bf Verifier-free reinforcement learning.}
Probability-based RL uses the model's own confidence in reference answers
to provide supervision without external verifiers.
JEPO~\citep{tang2025beyond} and NOVER~\citep{liu2025nover} construct
training signals from answer log-probabilities and token-level
perplexities, respectively. Rita~\citep{sun2026aligningthoughtsanswersprobability} leverages bounding box-based posterior probability in visual grounding tasks to regularize thinking-answer consistency of vision-language models. More directly related to our work,
VeriFree~\citep{zhou2025reinforcing} rewards reasoning traces using
posterior probabilities of reference answers, while
RLPR~\citep{yu2025rlpr} extends this approach to non-verifiable domains
through length normalization and reward debiasing. These methods
demonstrate the promise of probability-based supervision, but its
reliability for long reasoning traces remains underexplored.
We address this gap by analyzing posterior concentration and introducing
RLCPR, which mitigates its effects through uncertainty-aware data
sampling and concentration-aware regularization.

%% file: 2.5_preliminary.tex
\subsection{Posterior Probability-Based RL}\label{sec:probRL}

Probability-based RL methods construct rewards from the model’s conditional probabilities of the reference-answer tokens. It can be generally formulated as:
\begin{equation}\label{eq:posterior}
    R^\text{Post}_\theta(z)=\tilde{P}(Y=y^*\Cond Z=z; x;\theta),
\end{equation}
where $z$ is the reasoning  sequence generated by the LLM, $y^*$ is the ground-truth answer, $x$ is the input prompt, and $Z$ and $Y$ are the random variables of reasoning sequence and prediction sequence, respectively. Intuitively, \Cref{eq:posterior} represents the model’s posterior belief in the reference answer conditioned on a reasoning trace. Existing methods instantiate this posterior-derived signal through different token-level aggregations, \emph{i.e.} \(r=f(R_\theta(z))\) with different $f$, such as sequence likelihood in VeriFree~\citep{zhou2025reinforcing} and token-level probability mean in RLPR~\citep{yu2025rlpr}. 

Rewards such as \Cref{eq:posterior} are used in RL training with methods such as GRPO~\citep{grpo}. For each input \(q\), the old policy model samples \(N\) candidate outputs \(\{o_i\}_{i=1}^N\) to form a group. Their corresponding rewards \(\{r_i\}_{i=1}^N\) are then normalized within the group, \emph{i.e.} $A_i \;=\; \frac{r_i - \mu}{\sigma}$, to compute relative advantages.
It measures each candidate’s quality with respect to its siblings. The GRPO objective is
\begin{equation}
\mathcal{J}_{\mathrm{GRPO}}(\theta) = \mathbb{E}_{\{o_i\}\sim \pi_{\theta_{\text{old}}}(q)} \Bigg[
\frac{1}{N}\sum_{i=1}^N
\Big\{
\min\!\big[s_1 A_i,\; s_2 A_i\big]
 -  \beta\,\mathbb{D}_{\mathrm{KL}}\!\big[\pi_\theta \,\|\, \pi_{\mathrm{ref}}\big]
\Big\}
\Bigg],
\end{equation}
where $s_1=\frac{\pi_\theta(o_i\mid q)}{\pi_{\theta_{\text{old}}}(o_i\mid q)}$ and $s_2  =\operatorname{clip}\!\left(\frac{\pi_\theta(o_i\mid q)}{\pi_{\theta_{\text{old}}}(o_i\mid q)},\, 1-\epsilon,\, 1+\epsilon\right)$ \textcolor{black}{with the clipping function that limits the interval to $[1-\epsilon,1+\epsilon]$.}

To improve optimization efficacy, %
RLPR \citep{yu2025rlpr} filters %
low-variance groups.  Let $\sigma(g)$ denote the standard deviation of group $g$'s rewards; the group is retained only if
\begin{equation} 
\sigma(g) > \operatorname{EMA}_{j\in\mathcal G}[\sigma(j)]\times\beta, \label{eq:reward_filtering} 
\end{equation}
where $\mathcal{G}$ denotes all the sampled groups and $\text{EMA}_{j\in\mathcal{G}}$ is the exponential moving average of all the within-group standard deviations. As filtering is performed %
after the rollouts are generated and scored, low-information groups still incur the full generation cost. In \Cref{sec:3_theory}, we show that such low-variance posterior rewards can arise systematically from posterior concentration in long reasoning traces.

\subsection{Entropy and Log-Likelihood Concentration for LLMs}\label{sec:aep}
Let $\widetilde P(o_t\mid[x,o_{1:t-1}];\theta)$ denote the normalized autoregressive distribution induced by vocabulary truncation during decoding~\citep{holtzman2019curious,hewitt2022truncation}.
\cite{bell2024slaves} establish an un-equipartition property showing that the length-normalized NLL of a model-generated sequence converges in probability to its length-normalized realized entropy:
\begin{equation}
\lim_{N\rightarrow\infty}
\Pr\left(
\left|
-\frac{1}{N}\log\widetilde P(o_{1:N}\mid x;\theta)
-
\frac{1}{N}h_{\mathrm{realized}}(o_{1:N})
\right|>\epsilon
\right)
=0,
\label{eq:unequipartition}
\end{equation}
where $h_{\mathrm{realized}}(o_{1:N})
:=
\sum_{t=1}^{N}
\widetilde H
\bigl(
O_t \mid [x,o_{1:t-1}];\theta
\bigr)$ is the \textit{realized entropy} aggregating the next-token entropy along the trajectory. Its expectation is the deterministic full-sequence entropy:
\begin{equation}
\mathbb{E}_{o_{1:N}}
\left[
h_{\mathrm{realized}}(o_{1:N})
\right]
=
\widetilde H(O_{1:N}\mid x;\theta)
=
\sum_{t=1}^{N}h_{x,\theta}(t),
\label{eq:sequence_entropy}
\end{equation}
where
\(h_{x,\theta}(t)
=
\mathbb{E}_{o_{1:t-1}}
[
\widetilde H(O_t\mid[x,o_{1:t-1}];\theta)
]\).

Building on this connection, \cite{yang2026llmprobabilityconcentrationalignment} empirically show that long-sequence likelihoods can be characterized by the deterministic sequence entropy, motivating the approximation
\begin{equation}
-\frac{1}{N}\log\widetilde P(o_{1:N}\mid x;\theta)
\approx
\frac{1}{N}\widetilde H(O_{1:N}\mid x;\theta).
\label{eq:ll_concentration}
\end{equation}
Consequently, sufficiently long sequences of the same length tend to have similar NLLs despite following different generation trajectories. We refer to this tendency as \emph{log-likelihood concentration}.

\cite{yang2026llmprobabilityconcentrationalignment} further characterize the effective generative breadth through the \textit{branching factor}:
\begin{equation}
B_N(x;\theta)
:=
\exp\left(
\frac{1}{N}\widetilde H(O_{1:N}\mid x;\theta)
\right).
\label{eq:branching_factor}
\end{equation}
A larger value indicates a more diffuse generative distribution. Its empirically observed decrease with generation length suggests that later positions become increasingly predictable and motivates the entropy condition used in our PCP analysis.

%% file: 3_theory.tex
This section rethinks the concentration phenomenon of LLMs from the perspective of posterior probabilities. We are guided by two questions: \textit{Given two reasoning traces generated by the same model for the same prompt, does the posterior probability reflect their reasoning quality? If not, under what conditions does it become less informative for GRPO optimization?}  We explore with an empirical study %
in \Cref{sec:exploration}, before %
introducing our proposed Posterior Concentration Phenomenon (PCP) in \Cref{sec:general_pcp}. 

\subsection{Warm-up: Exploration Experiments}\label{sec:exploration}
\begin{figure}
    \centering
    \begin{subfigure}[t]{0.24\linewidth}
        \centering
        \includegraphics[width=\linewidth]{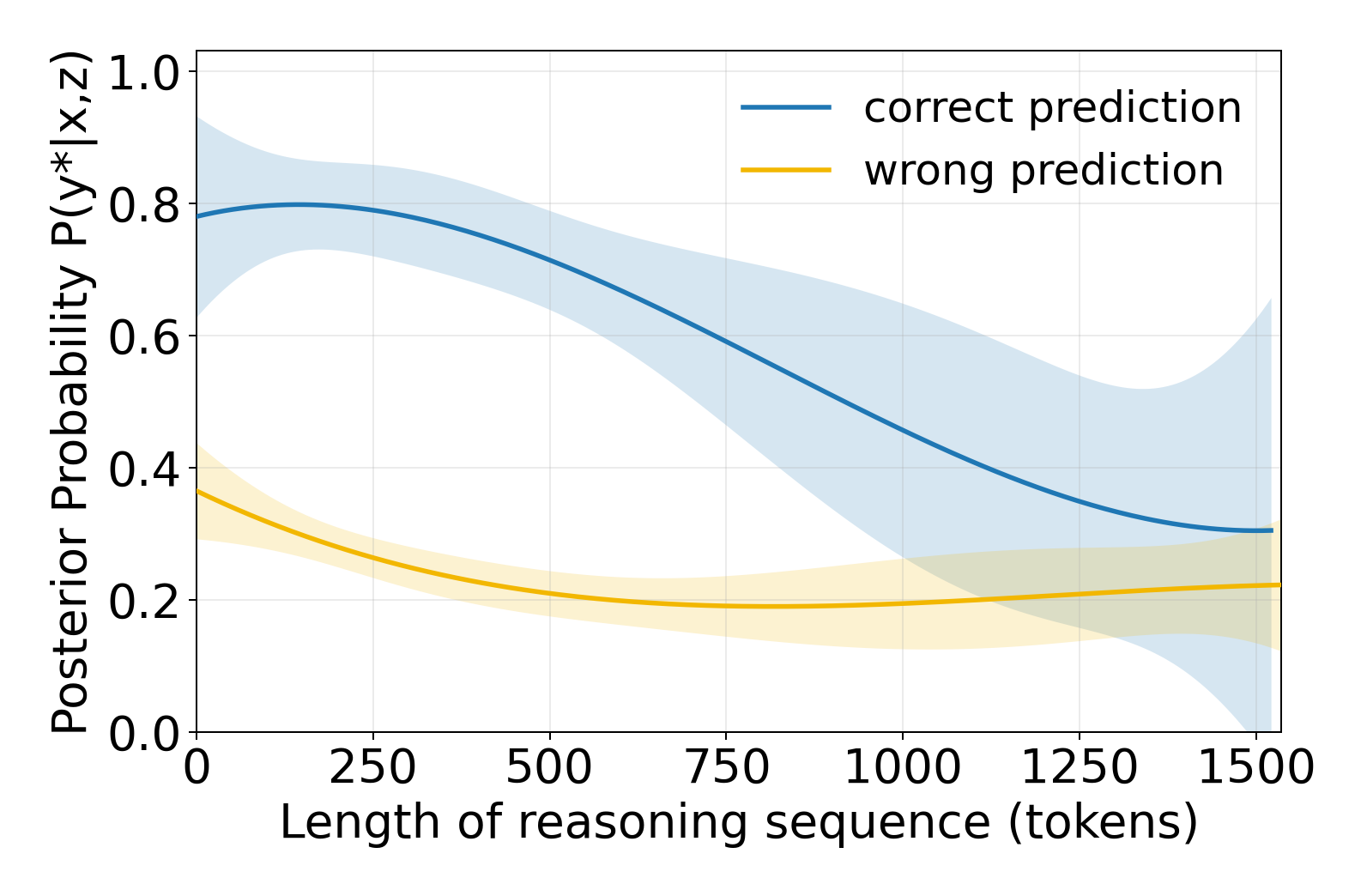}
        \caption{MMLU-Pro}
        \label{fig:mmlupro_posterior}
    \end{subfigure}
    \begin{subfigure}[t]{0.24\linewidth}
        \centering
        \includegraphics[width=\linewidth]{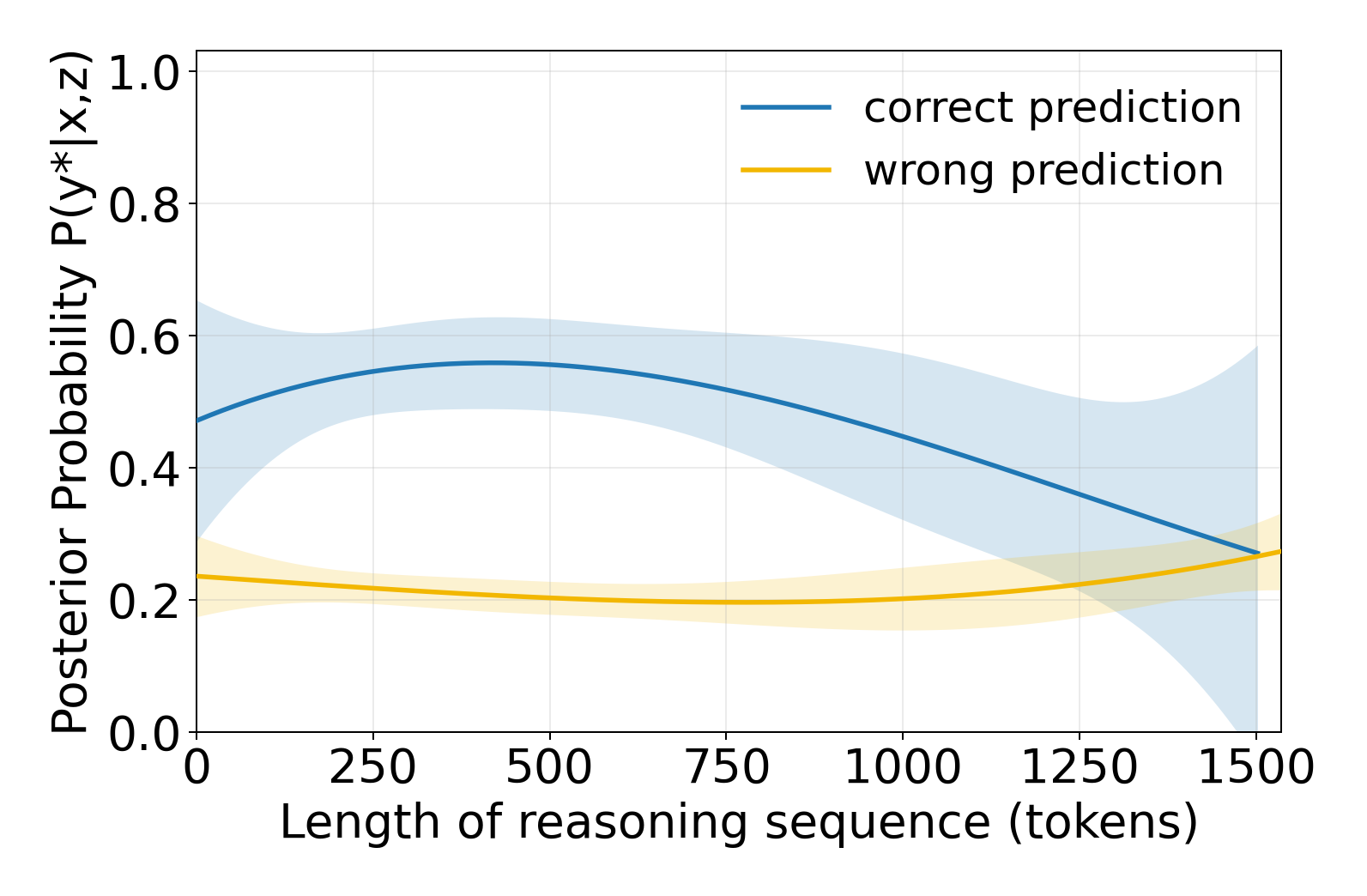}
        \caption{GPQA-Diamond}
        \label{fig:gpqa_posterior}
    \end{subfigure}
    \begin{subfigure}[t]{0.24\linewidth}
        \centering
        \includegraphics[width=\linewidth]{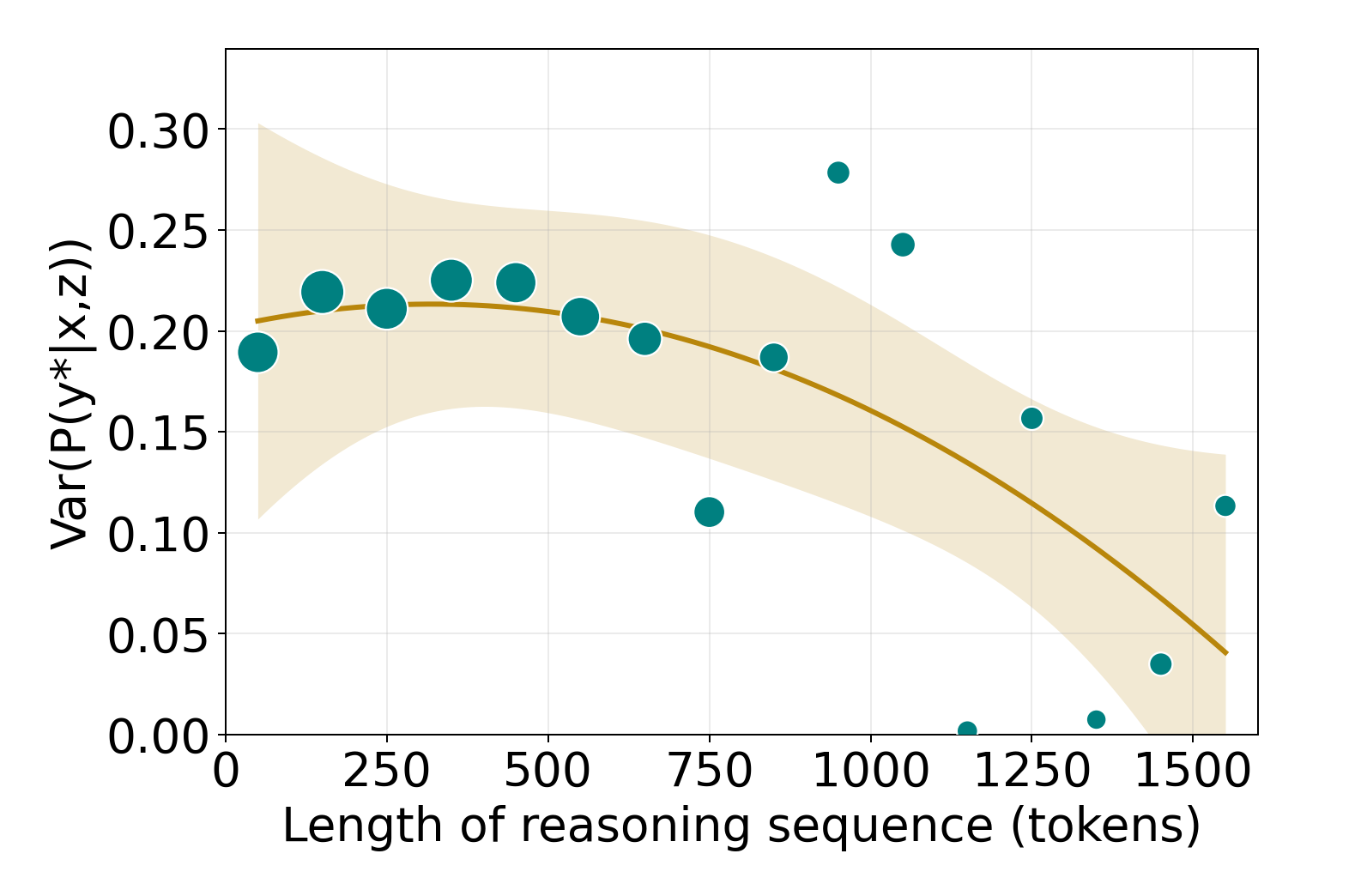}
        \caption{MMLU-Pro (Var.)}\label{fig:mmlupro_var}
    \end{subfigure}
    \begin{subfigure}[t]{0.24\linewidth}
        \centering
        \includegraphics[width=\linewidth]{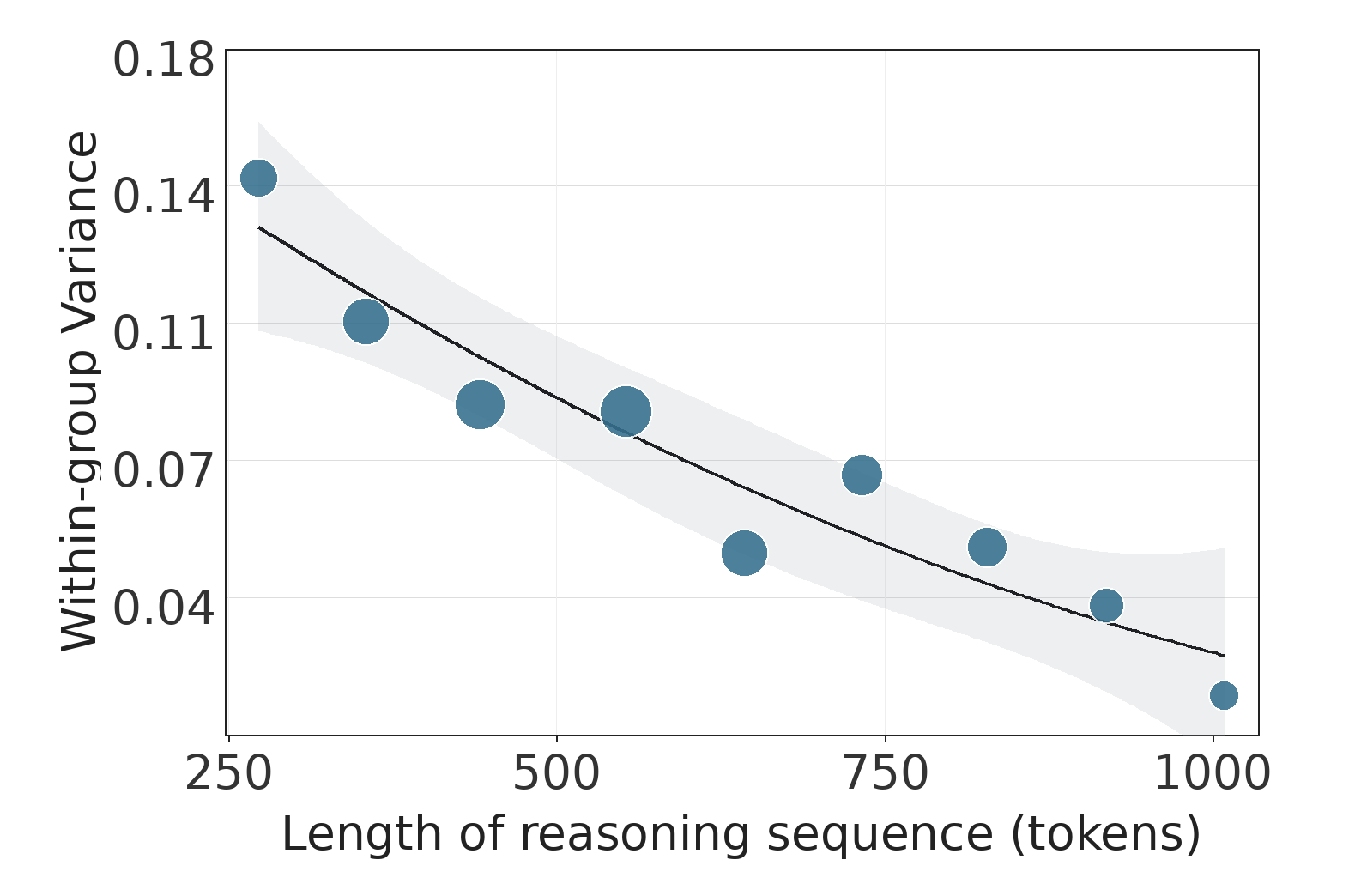}
        \captionsetup{
        justification=centering,
        font=footnotesize,
        skip=2pt
    }
        \caption{GPQA-Diamond \\[-1pt](Within-group Var.)}\label{fig:gpqa_var}
    \end{subfigure}
    \caption{\textbf{Posterior rewards become less discriminative as reasoning length grows.} On MMLU-Pro and GPQA-Diamond, posterior rewards for traces ending in \textcolor{Cyan}{correct} and \textcolor{Tan}{incorrect} answers approach one another at longer lengths (a, b). Also, the reward variance (c) and within-group variance (d) contract in the long-length regime. The shaded area represents the 95\% confidence interval. These experimental results point to an effect we named as \textbf{Posterior Concentration Phenomenon}, where posterior probability collapses to a similar value as the reasoning length grows.}
    \vspace{-0.14in}
    \label{fig:concentration_test}
\end{figure}
We begin by examining posterior rewards over progressively longer reasoning traces. \Cref{fig:concentration_test} shows two consistent signatures on the benchmarks MMLU-Pro~\citep{mmlupro} and GPQA-Diamond~\citep{gpqa}. First, posterior rewards for traces ending in correct versus incorrect answers become less distinguishable at longer lengths (\Cref{fig:mmlupro_posterior,fig:gpqa_posterior}, blue vs. yellow). 
Second, posterior-reward dispersion contracts with length: the overall reward variance decreases (\Cref{fig:mmlupro_var}), and the variance among rollouts within the same group also steadily declines (\Cref{fig:gpqa_var}). Together, these results show that long reasoning traces provide less discriminative rewards within the groups used for relative policy optimization. We refer to this length-dependent loss of discriminative power as the \textbf{Posterior Concentration Phenomenon (PCP)}. See Appx.~\ref{ap:exploration_setup} for the details of this exploration experiment.

\subsection{Posterior Concentration Phenomenon}\label{sec:general_pcp}
Let $Z$ and $Y$ denote %
{the random variables representing} the reasoning and answer sequences, %
{where the LLM first generates the reasoning
trace and then the answer}. For %
a reasoning trace $z$ and reference
answer $y^*$, the posterior reward can be separated into two terms:
\begin{equation}\label{eq:r_expansion}
\log \widetilde P(y^* \mid x,z;\theta)
=
\log \widetilde P(z,y^* \mid x;\theta)
-
\log \widetilde P(z \mid x;\theta), 
\end{equation}
{where $\widetilde P$ represents the normalized generative distribution.}
Let $n=|z|$ and $m=|y^*|$ {denote the lengths of the %
reasoning %
and answer sequences respectively}. 
We define the residual of log-likelihood concentration as $\epsilon_n(o_{1:n}):=-\log \widetilde P(o_{1:n}\mid x;\theta)-\widetilde H_n(x;\theta)$,
based on the approximation in \Cref{eq:ll_concentration}, 
where $\widetilde H_n(x;\theta)=\widetilde H(O_{1:n}\mid x;\theta)$ {denotes the expected %
entropy over a length-$n$ sequence}. %
{By substituting the residuals into %
\Cref{eq:r_expansion}, we obtain}
\begin{equation}\label{eq:log_prob_approx}
\log \widetilde P(y^* \mid x,z;\theta)
=
-\left[\widetilde H_{n+m}(x;\theta)-\widetilde H_n(x;\theta)\right]
+
\left[\epsilon_n(z)-\epsilon_{n+m}(z,y^*)\right].
\end{equation}
\Cref{eq:log_prob_approx} implies that a log-posterior probability, at a sequence-level, can be decomposed into two deterministic %
entropies and the corresponding residuals pushing them away from these approximations.  
Now, consider two %
{independent} traces $z^1,z^2$ with $|z^1|=n+s, |z^2|=n$, where $n$ is sufficiently large and $s$ is a small positive integer. %
{The log ratio between two posteriors can then be derived} from \Cref{eq:r_expansion,eq:log_prob_approx} {as}:
\begin{equation}\label{eq:gen_relative}
\resizebox{0.98\hsize}{!}{%
    $\log \frac{\widetilde{P}(y^*\Cond x, z^1;\theta)}{\widetilde{P}(y^*\Cond x, z^2;\theta)}=-\left[\widetilde{H}_{n+s+m}(x;\theta)-\widetilde{H}_{n+s}(x;\theta)\right]+\left[\widetilde{H}_{n+m}(x;\theta)-\widetilde{H}_{n}(x;\theta)\right]+\rho_{n+s}(z^1)-\rho_n(z^2),$
    }
\end{equation}
where $\rho_n(z)=\epsilon_n(z)-\epsilon_{n+m}(z,y^*)$ is the length-induced residual shift that measures the differences between log-likelihood concentrations {before and after} generating the answer $y^*$. 
Recall from \Cref{eq:sequence_entropy} that $\widetilde{H}_{n}(x;\theta)=\sum_{t=1}^n h_{x,\theta}(t)$.
If we define the answer-window entropy after a length-$n$ reasoning trace as $W_n=\widetilde{H}_{n+m}(x;\theta)-\widetilde{H}_{n}(x;\theta)=\sum_{t=n+1}^{n+m}h_{x,\theta}(t)$, then \Cref{eq:gen_relative} simplifies into:
\begin{equation}\label{eq:structured_gen_relative}
    \log \frac{\widetilde{P}(y^*\Cond x, z^1;\theta)}{\widetilde{P}(y^*\Cond x, z^2;\theta)}=\underbrace{W_n-W_{n+s}}_{\text{length-induced entropy shift}}+\underbrace{\rho_{n+s}(z^1)-\rho_n(z^2)}_{\text{concentration residuals}}.
\end{equation}

\Cref{eq:structured_gen_relative} decomposes the unequal-length posterior difference into two parts.
The first term, $W_n-W_{n+s}$, is a deterministic, length-induced entropy
shift.  The second term,
$\rho_{n+s}(z^1)-\rho_n(z^2)$, captures the trace-specific residuals of the
log-likelihood approximation. 
For equal-length traces, \emph{i.e.} $s=0$, the deterministic entropy shift cancels exactly, and the log-posterior difference depends entirely on the trace-specific residual difference. For unequal-length traces, we additionally characterize the entropy shift $W_n-W_{n+s}$ by the following lemma.

Define the length-averaged token entropy as
$\bar h_{x,\theta}(n):=\widetilde H_n(x;\theta)/n
=\frac{1}{n}\sum_{t=1}^{n}h_{x,\theta}(t)$.
Notably, $\exp(\bar h_{x,\theta}(n))$ is exactly the branching factor
defined in \Cref{eq:branching_factor}. Prior work empirically observes that
this quantity generally decreases as generation progresses
\citep{yang2026llmprobabilityconcentrationalignment}.
Motivated by this, we consider eventual non-increase of
$\bar h_{x,\theta}(n)$ as a sufficient condition and establish the following
lemma, with its proof deferred to
Appx.~\ref{ap:lemma_proof}.
\begin{lemma}[Asymptotic Answer-Window Stability]
\label{entropy_stability_lemma}
If $\bar h_{x,\theta}(n)$ is eventually non-increasing, then for any fixed
positive integers $m$ and $s$, and any $\epsilon>0$,
\begin{equation} \lim_{N\to\infty} \frac{1}{N} \Big| \Big\{ n\in\{1,\ldots,N\}: \left|W_{n+s}-W_n\right|>\epsilon \Big\} \Big| =0. \label{eq:answer_window_density} \end{equation}
\end{lemma}

Intuitively, \Cref{entropy_stability_lemma} states that, as the length range grows, large differences in answer-window entropy with a length gap $s$ becomes more rare.
Notably, this does not require
the token-level entropy to vary smoothly. Local fluctuations with particularly high-entropy tokens may persist, but non-negligible shifts on answer-window entropies become
asymptotically sparse. 
Combining \Cref{entropy_stability_lemma} with \Cref{eq:structured_gen_relative}, we obtain the following PCP characterization:
\begin{proposition}[Posterior Concentration Phenomenon.]
\label{general_prop}
Let $Z_n$ denote the random reasoning sequence generated by the model conditioned on having length $n$. Let
$L_N\sim\mathrm{Unif}({1,\ldots,N})$ be an auxiliary random reasoning
length. Suppose that $\bar h_{x,\theta}(n)$ is eventually non-increasing.
Given a fixed $s>0$, if, for any $\delta>0$,
$\mathbb{P}\left(
\left|
\rho_{n+s}(Z_{n+s})-\rho_n(Z_n)
\right|>\delta
\right)
\to0
\text{ as }n\to\infty,
$
then, for any $\epsilon>0$,
\begin{equation}
\lim_{N\to\infty}
\mathbb{P}
\left(
\left|
\log
\frac{
\widetilde P(y^* \Cond x,Z_{L_N+s};\theta)
}{
\widetilde P(y^* \Cond x,Z_{L_N};\theta)
}
\right|>\epsilon
\right)
=0,
\end{equation}
where the probability is taken jointly over the auxiliary length $L_N$
and the random reasoning sequences $Z_{L_N+s}$ and
$Z_{L_N}$.  See proof in Appx.~\ref{ap:proof_general_pcp}.
\end{proposition}

Above, $L_N$ serves as a probabilistic counterpart to the deterministic length index $n$ in \Cref{entropy_stability_lemma}: uniformly sampling $L_N$ converts \Cref{entropy_stability_lemma}'s zero-density statement over reasoning lengths into an equivalent probability statement.
Intuitively, \Cref{general_prop} suggests that, as the considered range of
reasoning length 
grows, posterior probabilities for independently sampled traces of similar lengths become %
more likely to take similar values when their residual differences are close. 
Note that the sufficient condition of residual difference is motivated by the empirical observations from \cite{yang2026llmprobabilityconcentrationalignment}. See Appx.~\ref{ap:discussion} for further discussion.
Importantly, this does not imply lower model confidence; rather, the
posterior becomes less discriminative about the quality of the underlying
reasoning trace. This observation complements recent findings that
reasoning can distort probability-based confidence signals~\citep{welch2026cost,fu2025multiple,xiong2026cora}. 
It uncovers the limitation of answer posterior probabilities as supervision signals:
internal confidence or consistency does not necessarily provide reliable
discrimination of reasoning quality.
From an RL perspective, this loss of discriminability weakens the learning
signal provided by posterior rewards: long rollouts may receive similar
rewards despite differences in their reasoning quality, potentially making the relative advantages less informative about reasoning quality while still incurring substantial
generation cost. {Since PCP is a probabilistic effect,} this motivates a PCP-aware framework with two complementary
directions: (1) preferentially sample training examples that are less prone to
posterior concentration, and (2) discourage unnecessarily long reasoning once
posterior difference collapses. We introduce these two components in
\Cref{sec:4_method}.

%% file: 4_method.tex
\begin{figure}
    \centering
    \includegraphics[width=\linewidth]{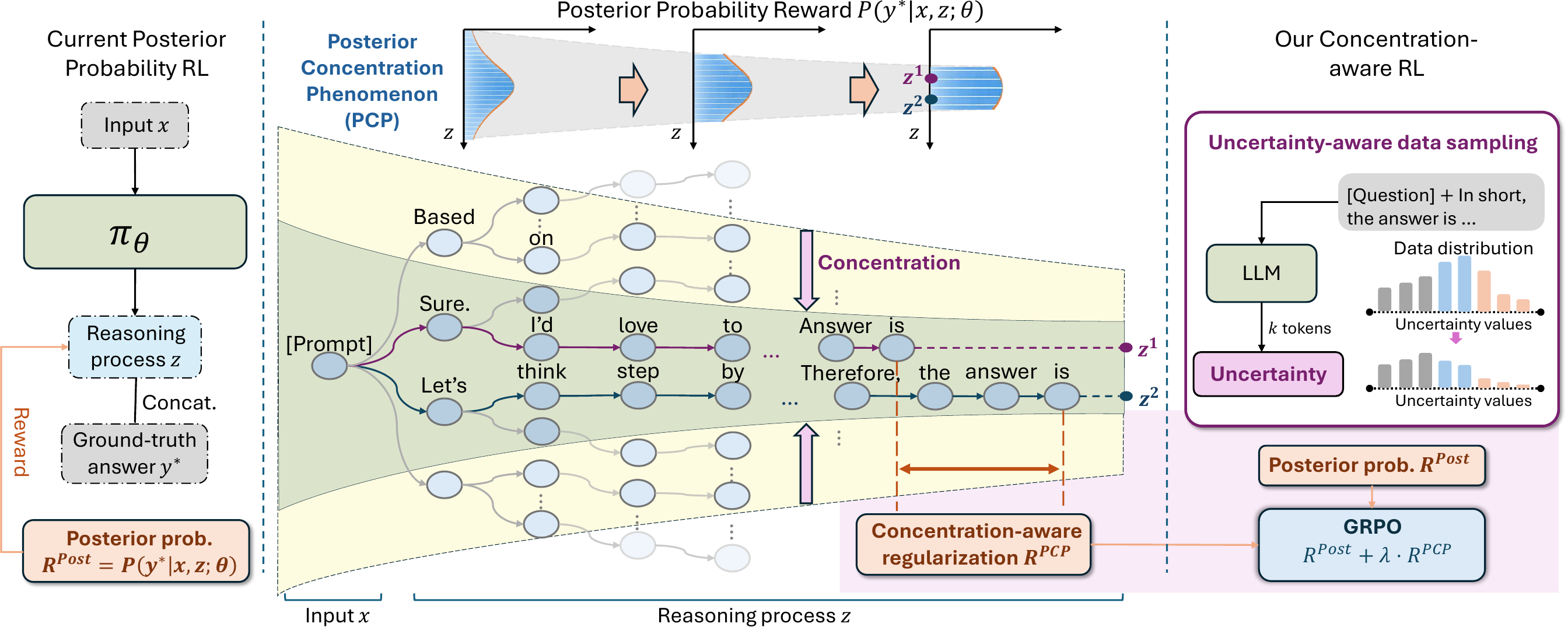}
    \caption{\textbf{Overview of RLCPR.} {RLCPR is a PCP-aware training framework with two components: Uncertainty-aware data sampling (UDS) reshapes the data distribution according to direct-answer uncertainty, reducing rollouts that are more prone to posterior concentration; Concentration-aware regularization (CR) detects concentrated rollout groups and introduces a length-aware penalty to discourage unnecessarily long reasoning traces.}}
    \vspace{-0.15in}
    \label{fig:overview}
\end{figure}
To account for posterior concentration during policy optimization, 
we propose RLCPR, a novel verifier-free post-training framework with two core components: uncertainty-aware data sampling (UDS) and concentration-aware regularization (CR). 

\noindent{\bf Uncertainty-aware data sampling (UDS).} %
{Discarding less informative groups with low standard deviation, as done by \cite{yu2025rlpr}, is inefficient. %
We propose to use uncertainty-aware data sampling to reduce unnecessary rollouts that may potentially suffer from PCP. 
Our sampling strategy is motivated by an empirical association:
questions whose rollout groups exhibit posterior concentration
tend to have higher direct-answer uncertainty during training.
This suggests using uncertainty as a proxy for concentration
risk and preferentially retaining lower-uncertainty examples.
Appx.~\ref{ap:uncertainty_concentration} presents the
diagnostic experiment about this relation between uncertainty and concentration.

Specifically, for each question $x$, we {concatenate a suffix $q$}, and define the model's uncertainty 
as the average token entropy of the $k$ tokens after the sequence $[x, q]$.
{In this paper, our} suffix $q$ is \textit{``In short, the answer is ''}. This uncertainty can be formulated as:
\begin{equation}\label{eq:uncertainty}
    U_\theta(x)=\frac{1}{k}\sum_{t=1}^k \tilde{H}\big(O_t \Cond [x, q, \hat{o}_{1:t-1}];\theta\big),
\end{equation}
where 
$\hat{o}_{1:t-1}$ is the output sequence until position $t-1$ under greeding decoding strategy, \emph{i.e.}, $\hat{o}_t=\argmax_{v\in V}\tilde{P}\left(v\Cond [x, q, \hat{o}_{1:t-1}] ; \theta\right)$ with $V$ as the vocabulary. We refer to this quantity as \textit{direct-answer uncertainty}. During the experiments, we use a buffer storage to save the uncertainty values and update this buffer every $f$ GRPO steps. This process is equivalent to using fixed model parameters $\hat{\theta}$ to estimate the uncertainty $U_\theta(x)$ with $U_{\hat{\theta}}(x)$ and periodically updating $\hat{\theta}$ after every $f$ steps. It achieves high computational efficiency while preserving reliable approximation to the model's uncertainty to different question-response pairs.

Based on the proposed uncertainty measure, we introduce two quantile-based thresholds $\tau_1$ and $\tau_2$ with $\tau_0=0\leq\tau_1\leq\tau_2\leq1=\tau_3$ to divide the question batch into three bins $b_1,b_2,b_3$. More precisely, this process can be represented with the following expression:
\begin{equation}\label{eq:bin_division}
    \mathbbm{1}\Big(x\in b_i\Big)\equiv\mathbbm{1}\Big(\tau_{i-1}\leq P_{x'\sim\mathcal{D}}\big(U_\theta(x)>U_\theta(x')\big)<\tau_i\Big),
\end{equation}
{where $\mathbbm{1}(\cdot)$ is the binary indicator function.} Then, we use the filtering rates $\omega_1,\omega_2,\omega_3$ to randomly remove the data in $b_1,b_2,b_3$. This operation is equivalent to modifying the sampling strategy such that the data sampling distribution is related to the model uncertainty level. Unlike post-generation reward filtering, UDS filters questions
before generating full reasoning traces. Its uncertainty
estimates require only short direct-answer completions and
are cached across training steps, amortizing the estimation
cost. Appx.~\ref{ap:complexity} provides the
complexity analysis.

\noindent{\bf Concentration-aware regularization (CR).}
We also propose a concentration-aware regularizer for GRPO-based training. When concentration happens, the posterior-based rewards %
no longer reflect the quality of reasoning processes,  %
so we argue that the model should learn to reduce the reasoning lengths outright. %
Inspired by \cite{team2025kimi}, we adopt a length penalty to penalize relatively longer reasoning sequences and encourage relatively shorter reasoning sequences within a group. Different from \cite{team2025kimi}, our length penalty is used only when concentration happens. Specifically, consider %
a %
gap of posterior probabilities for a group $G$ as follows:
\begin{equation}\label{eq:gap}
    \Delta G=\max\limits_{z^1,z^2\in G} \left|\tilde{P}(y^*_G\Cond x_G,z^1;\theta) - \tilde{P}(y^*_G\Cond x_G,z^2;\theta)\right|,
\end{equation}
where $x_G,y^*_G$ are the input question and reference answer corresponding to group $G$. We then propose the following penalty:
\begin{equation}\label{eq:length_penalty}
    R_\theta^{\text{PCP}}(z)=-\mathbbm{1}\Big(l_{\text{min}}>\delta\;\text{and}\;\Delta G_z<\alpha\Big)
    \cdot\frac{\left|z\right|-l_{\text{min}}}{\max\{l_{\text{max}}-l_{\text{min}},1\}},
\end{equation}
where $G_z$ denotes the 
{group corresponding to $z$} in GRPO, $l_{\text{min}}=\min\limits_{z'\in G_z} |z'|$, and $l_{\text{max}}=\max\limits_{z'\in G_z} |z'|$.
Intuitively, \Cref{eq:length_penalty} defines a binary indicator to detect the groups with PCP and imposes the length-aware penalty on these groups to encourage shorter responses. %
Concentration occurs %
if all the reasoning sequences are sufficiently long (with respect to $\delta$) and the posterior probabilities are %
similar %
(with respect to $\alpha$). The two thresholds $\alpha$ and $\delta$ are decided at runtime. 
{Specifically, we compute quantiles over the training batch to adapt to the variation of policy models and data distributions:}
\begin{equation}
\delta
=
Q_{p_\delta}\!\left(\left\{|z|:z\in G,\;G\in\mathcal{B}\right\}\right),
\qquad
\alpha
=
Q_{p_\alpha}\!\left(\left\{\Delta G:G\in\mathcal{B}\right\}\right),
\label{eq:dynamic_thresholds}
\end{equation}
where $\mathcal{B}$ denotes the rollout groups in the current training batch and $Q_p(\cdot)$ denotes the empirical $p$-quantile.
The final concentration-aware (CA) reward for a given rollout is then defined as the weighted sum of the posterior-based reward and the PCP-aware regularization, \emph{i.e.}
\begin{equation}\label{eq:weighted_sum}
    R^\text{CA}_\theta(z) = R^\text{Post}_\theta(z)+\lambda R^\text{PCP}_\theta(z).
\end{equation}
\Cref{eq:weighted_sum} suggests a straightforward learning objective: When little information can be extracted solely from the posterior probability, the model should learn to shorten the length.
Building on our empirical experience, we suggest a cold start for the PCP-aware penalty, i.e. a lower $\lambda$ at the beginning and gradually increasing the weight.

%% file: 5_experiments.tex
\input{tab/quantitative}
\vspace{-0.4em}
\subsection{Experimental settings}
\vspace{-0.3em}
\noindent{\bf Model and implementation.} We evaluate RLCPR on Qwen3-4B-Base~\citep{yang2025qwen3}. Our implementation builds on the official RLPR~\citep{yu2025rlpr} codebase in \texttt{verl}~\citep{verl} {but we exclude the standard-deviation filtering}. For UDS, we set $k=3$, $f=20$, and $(\tau_1,\tau_2)=(0.3,0.7)$ with linear annealing during the first 200 steps; the filtering rates are $(\omega_1,\omega_2,\omega_3)=(0.2,0.45,0.7)$, where data with higher uncertainty are associated with lower sampling rates. For CR, $p_\delta$ and $p_\alpha$ are set to 0.8 and 0.2, respectively. We train for 600 steps to achieve an output-token budget comparable to RLPR, with $\lambda=0$ for the first 160 steps and $0.5$ thereafter. Other hyperparameters follow RLPR. Implementation details and training prompt can be found in Appx.~\ref{ap:main_details} and~\ref{ap:prompt}.

\vspace{-0.3em}
\noindent{\bf Baselines.} 
Our primary baseline is RLPR with the same backbone. We additionally report the Qwen2.5-7B results excerpted from \cite{yu2025rlpr}, including RLVR~\citep{hu2026open}, General Reasoner~\citep{general-reasoner}, VeriFree~\citep{zhou2025reinforcing}, and RLPR~\citep{yu2025rlpr}.

\vspace{-0.3em}
\noindent{\bf Datasets.} 
Following RLPR, we train on 77K non-mathematical reasoning prompts from ~\cite{general-reasoner}, filtered by GPT-4.1~\citep{gpt4-1}. %
We evaluate on four general-domain benchmarks--MMLU-Pro~\citep{mmlupro}, GPQA-Diamond~\citep{gpqa}, TheoremQA~\citep{theoremqa}, and the held-out WebInstruct subset~\citep{general-reasoner}; and three mathematical benchmarks--MATH-500~\citep{math500}, Minerva~\citep{minerva}, and AIME24~\citep{aime24}.
\vspace{-0.2em}
\subsection{Main results}\label{subsec:main_results}
\vspace{-0.2em}
\Cref{tab:quan} summarizes performance on seven reasoning benchmarks. On Qwen3-4B-Base, RLCPR improves average accuracy from \(56.2\%\) to \(58.0\%\), outperforming RLPR on six of seven benchmarks. It gains 2.8, 4.0, and 0.9 percentage points on MMLU-Pro, TheoremQA, and WebInstruct, respectively. Despite training exclusively on non-mathematical prompts, RLCPR also improves MATH-500, Minerva, and AIME24 by 2.2, 3.0, and 2.3 points, demonstrating benefits beyond the training domain.

RLCPR also improves rollout-token efficiency. As shown in \Cref{fig:efficiency}, at 320 steps, RLCPR uses 0.75 billion rollout tokens, \(62.3\%\) fewer than RLPR's 1.99 billion, while still achieving higher accuracy; even after 600 steps, it uses only 1.66 billion tokens. See Appx.~\ref{ap:complexity} for the complexity analysis. RLCPR nevertheless shows a 2.7\% decrease on GPQA-Diamond relative to RLPR. This may partly arise because favoring lower-uncertainty examples and shorter traces could limit learning opportunities for challenging problems requiring extended reasoning. Although this explanation warrants further investigation, the benchmark-specific decrease does not alter the overall gains in average accuracy and rollout-token efficiency.

\subsection{Ablation Study}\label{subsec:ablation}

\input{tab/ablation}
\noindent{\bf UDS and CR effect.} \Cref{tab:component_ablation} evaluates the individual and joint effects of UDS and CR. CR alone provides modest gains, whereas UDS alone slightly degrades performance on MATH-500 and Minerva, possibly because it filters out some difficult but informative training instances. Importantly, UDS is not a static difficulty-based filter: it periodically re-estimates uncertainty under the current policy (every \(f=20\) steps) and dynamically reshapes the training distribution. Its effectiveness therefore depends on how the policy evolves. Without CR, the retained rollout groups may still suffer from PCP, causing their posterior rewards to concentrate and weakening the optimization signal that drives subsequent policy updates. CR complements UDS by augmenting concentrated posterior rewards with a length-aware penalty, thereby restoring a discriminative training signal within the affected groups and enabling the policy to learn more effectively from the UDS-selected samples. The improved policy can, in turn, yield more informative uncertainty estimates for subsequent UDS updates. This dynamic interaction may explain why combining UDS with CR achieves the best performance on all three benchmarks, improving over the baseline by up to \(6.7\%\).

\noindent{\bf CR cold start and UDS scheduler}
\Cref{tab:schedule_ablation} evaluates the CR
cold start and the UDS annealing scheduler.
Without the cold start, we apply CR with
$(\lambda,p_\delta,p_\alpha)=(0.5,0.8,0.2)$ from the
beginning of training. This reduces accuracy by 4.9,
8.6, and 7.9 percentage points on MMLU-Pro,
TheoremQA, and WebInstruct, respectively, suggesting
that early regularization can hinder learning.
Removing the UDS scheduler and fixing
$(\tau_1,\tau_2)=(0.3,0.7)$ throughout training
also lowers accuracy across all three benchmarks.
Together, these results support our claim that concentration becomes increasingly significant as the training proceeds. Thus, delaying CR and
gradually adjusting the UDS thresholds during
early training provides a more effective way to mitigate concentration.

%% file: tab/quantitative.tex
\begin{table}[t]
    \centering
    
    \caption{\textbf{Overall performance on seven reasoning benchmarks.}  $\mathrm{Avg}@k$ denotes accuracy estimated using $k$ sampled responses per question.
Compared with same-backbone RLPR, RLCPR outperforms RLPR on six of seven benchmarks and improves average accuracy by 1.8\% while achieving higher training token efficiency. WebInst. denotes the held-out evaluation subset of WebInstruct.}
\vspace{-0.5em}
    \resizebox{\columnwidth}{!}{%
    \renewcommand{\arraystretch}{1.1}
    \begin{tabular}{l|l|c|ccccccc|c}
    \thicktoprule
        \multirow{2}{*}{\textbf{Method}}&\multirow{2}{*}{\textbf{Backbone}}&\multirow{2}{*}{\textbf{Verifier}}&\textbf{MMLU-Pro}&\textbf{GPQA-Dia.}&\textbf{TheoremQA}&\textbf{WebInst.}&\textbf{MATH-500}&\textbf{Minerva}&\textbf{AIME24}&\multirow{2}{*}{\textbf{All}}\\
        &&&Avg@2&Avg@4&Avg@2&Avg@2&Avg@2&Avg@2&Avg@16&\\
        \specialrule{1.5pt}{0pt}{0pt}
        Base & \multirow{5}{*}{Qwen2.5-7B} & - & 45.3&32.4&41.4&60.4&63.0&37.6&6.5&40.9\\
        RLVR &  &Rule&55.1&36.2&52.2&75.3&76.5&54.9&\textbf{17.7}&52.6\\
        Gen. Reasoner &  &Model&55.4&37.4&52.1&74.5&77.0&51.7&16.0&52.0\\
        VeriFree&    &\cellcolor{JungleGreen!20}Free&53.8&36.7&47.6&72.5&73.5&49.0&12.5&49.4\\
        RLPR & &\cellcolor{JungleGreen!20}Free&56.0&37.6&55.4&75.5&78.0&56.5&16.3&53.6\\
        \arrayrulecolor{black!45}\hline %
\arrayrulecolor{black}          %
        Base  &&-&41.7&25.1&35.9&54.7&48.5&34.0&5.0&35.0\\
        RLPR &Qwen3-4B-Base&\cellcolor{JungleGreen!20}Free&58.5 &\textbf{43.4}&57.1&78.0&80.7&61.5&14.0&56.2\\
        \rowcolor{orange!25}
RLCPR
& \cellcolor{white}
& \cellcolor{JungleGreen!20}Free
& \textbf{61.3}\gain{2.8}
& 40.7
& \textbf{61.1}\gain{4.0}
& \textbf{78.9}\gain{0.9}
& \textbf{82.9}\gain{2.2}
& \textbf{64.5}\gain{3.0}
& {16.3}\gain{2.3}
& \textbf{58.0}\gain{1.8} \\
        \thickbottomrule
    \end{tabular}%
    }
    \label{tab:quan}
\end{table}

%% file: tab/ablation.tex
\begin{figure}[t]
    \centering
    \captionsetup{skip=3pt}

    \begin{minipage}[t]{0.5\linewidth}
        \vspace{0pt}
        \centering
        \includegraphics[width=\linewidth]{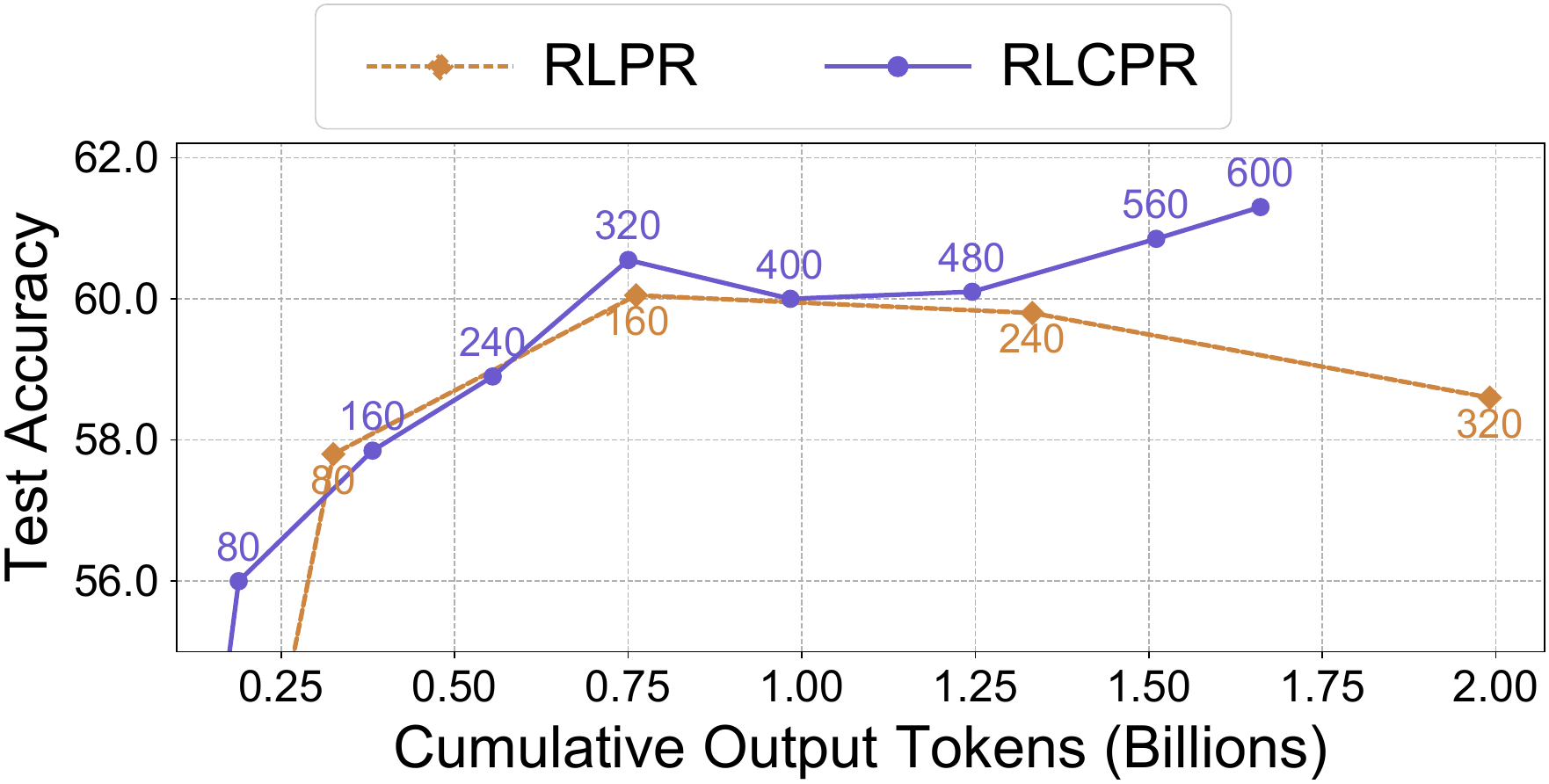}
        \caption{\textbf{Rollout-token efficiency.}
RLCPR achieves higher accuracy on MMLU-Pro with fewer cumulative output tokens. Numbers on the curves denote training steps.}
        \label{fig:efficiency}
    \end{minipage}
    \hfill
    \begin{minipage}[t]{0.48\linewidth}
        \vspace{0pt}
        \centering
        \footnotesize
        \renewcommand{\arraystretch}{1.12}
        \setlength{\tabcolsep}{2pt}
        \vspace{0.05in}
        \captionof{table}{Ablation of UDS and CR.}
    \label{tab:component_ablation}
    \centering
    \footnotesize
    \resizebox{.8\linewidth}{!}{%
        \begin{tabular}{@{}lccc@{}}
            \thicktoprule
            \textbf{Method}
                & \textbf{MATH-500}
                & \textbf{Minerva}
                & \textbf{AIME24} \\
            \hline
            Baseline    & 78.9 & 57.8 & 14.4 \\
            \quad + CR  & 79.5 & 59.7 & 15.0 \\
            \quad + UDS & 76.9 & 56.6 & 14.6 \\
            \hline
            RLCPR & \textbf{82.9}
                  & \textbf{64.5}
                  & \textbf{16.3} \\
            \thickbottomrule
        \end{tabular}%
    }

        \vspace{0.1in}
\captionof{table}{Ablation of CR cold start and UDS linear-annealing scheduler.}
    \label{tab:schedule_ablation}
    \centering
    \footnotesize
    \resizebox{\linewidth}{!}{%
        \begin{tabular}{@{}lccc@{}}
            \thicktoprule
            \textbf{Method}
                & \textbf{MMLU-Pro}
                & \textbf{TheoremQA}
                & \textbf{WebInstruct} \\
            \hline
            w/o CR cold start & 56.4 & 52.5 & 71.0 \\
            w/o UDS scheduler & 60.2 & 59.1 & 78.2 \\
            \hline
            RLCPR & \textbf{61.3}
                  & \textbf{61.1}
                  & \textbf{78.9} \\
            \thickbottomrule
        \end{tabular}%
    }
    \end{minipage}
    \vspace{-0.15in}
\end{figure}

%% file: 6_conclusion.tex
We investigate the reliability of posterior probability rewards for verifier-free RL and identify the Posterior Concentration Phenomenon (PCP), where rewards become less discriminative as reasoning traces grow longer. We characterize PCP empirically and provide a theoretical formulation connecting it to log-likelihood concentration. Guided by these findings, we propose RLCPR, which explicitly accounts for concentration through uncertainty-aware data sampling and selective length regularization. Experiments show that RLCPR outperforms RLPR on six of seven reasoning benchmarks while requiring fewer cumulative rollout tokens. These results highlight the value of concentration-aware training for improving both effectiveness and token efficiency in probability-based RL.

%% file: appendix.tex
\renewcommand{\thefigure}{A\arabic{figure}}
\renewcommand{\thetable}{A\arabic{table}}
\renewcommand{\theequation}{A\arabic{equation}}
\setcounter{figure}{0}
\setcounter{table}{0}
\setcounter{equation}{0}

\section*{Appendix}

\section{Proof of \Cref{entropy_stability_lemma}}
\label{ap:lemma_proof}

Since token-level entropy is non-negative,
$\bar h_{x,\theta}(n)\geq 0$. By assumption,
$\bar h_{x,\theta}(n)$ is eventually non-increasing, and therefore
converges to some finite limit
\begin{equation}
    \bar h_{x,\theta}(n)\to \bar h_{x,\theta}^{\infty}\geq 0.
    \label{eq:avg_entropy_limit}
\end{equation}
A finite prefix does not affect the asymptotic result, so in the
following we consider sufficiently large $n$ such that
$\bar h_{x,\theta}(n)$ is non-increasing.

By definition,
\begin{equation}
\begin{aligned}
    h_{x,\theta}(n)
    &=
    n\bar h_{x,\theta}(n)
    -(n-1)\bar h_{x,\theta}(n-1) \\
    &=
    \bar h_{x,\theta}(n)
    +(n-1)\bigl(
        \bar h_{x,\theta}(n)
        -\bar h_{x,\theta}(n-1)
    \bigr).
\end{aligned}
\label{eq:token_from_avg_entropy}
\end{equation}
Hence,
\begin{equation}
    0\leq h_{x,\theta}(n)
    \leq \bar h_{x,\theta}(n)
\end{equation}
for all sufficiently large $n$.

We next show that the token-level entropy approaches
$\bar h_{x,\theta}^{\infty}$ in the Ces\`aro $L^1$ sense. Since
$h_{x,\theta}(n)\leq\bar h_{x,\theta}(n)$,
\begin{equation}
\begin{aligned}
    &\frac{1}{N}\sum_{n=1}^{N}
    \left(
        \bar h_{x,\theta}(n)-h_{x,\theta}(n)
    \right) \\
    &\quad=
    \frac{1}{N}\sum_{n=1}^{N}\bar h_{x,\theta}(n)
    -\bar h_{x,\theta}(N)
    \to 0,
\end{aligned}
\label{eq:avg_entropy_gap}
\end{equation}
where the last step follows from
$\bar h_{x,\theta}(n)\to\bar h_{x,\theta}^{\infty}$ and Ces\`aro
convergence. Moreover,
$|\bar h_{x,\theta}(n)-\bar h_{x,\theta}^{\infty}|\to0$.
Therefore,
\begin{equation}
    \left|
        h_{x,\theta}(n)-\bar h_{x,\theta}^{\infty}
    \right|
    \leq
    \bar h_{x,\theta}(n)-h_{x,\theta}(n) 
    +
    \left|
        \bar h_{x,\theta}(n)-\bar h_{x,\theta}^{\infty}
    \right|,
\end{equation}
which gives
\begin{equation}
    \lim_{N\to\infty}
    \frac{1}{N}\sum_{n=1}^{N}
    \left|
        h_{x,\theta}(n)-\bar h_{x,\theta}^{\infty}
    \right|
    =0.
    \label{eq:token_entropy_cesaro}
\end{equation}

Recall that
$W_n=\sum_{j=1}^{m}h_{x,\theta}(n+j)$. For fixed $m$ and $s$,
\begin{equation}
\begin{aligned}
    |W_{n+s}-W_n|
    &\leq
    \sum_{j=1}^{m}
    \left|
        h_{x,\theta}(n+s+j)-h_{x,\theta}(n+j)
    \right| \\
    &\leq
    \sum_{j=1}^{m}
    \Bigl(
        \left|
            h_{x,\theta}(n+s+j)
            -\bar h_{x,\theta}^{\infty}
        \right| 
    +
        \left|
            h_{x,\theta}(n+j)
            -\bar h_{x,\theta}^{\infty}
        \right|
    \Bigr).
\end{aligned}
\label{eq:w_shift_bound}
\end{equation}
A fixed index shift preserves the Ces\`aro limit in
\Cref{eq:token_entropy_cesaro}. Thus,
\begin{equation}
    \lim_{N\to\infty}
    \frac{1}{N}\sum_{n=1}^{N}
    |W_{n+s}-W_n|
    =0.
    \label{eq:w_cesaro}
\end{equation}

Finally, consider an auxiliary random index
$n\sim\mathrm{Unif}({1,\ldots,N})$. By Markov's inequality,
\begin{equation}
\mathbb{P}_{n\sim\mathrm{Unif}({1,\ldots,N})}
\left(
|W_{n+s}-W_n|>\epsilon
\right) 
\leq
\frac{1}{\epsilon N}
\sum_{n=1}^{N}|W_{n+s}-W_n|
\to 0,
\end{equation}
where the last step follows from \Cref{eq:w_cesaro}.
The uniform distribution here is only an auxiliary device for measuring
the fraction of reasoning lengths that violate the stability condition.
Equivalently,
\begin{equation}
\begin{aligned}
\frac{1}{N}
\Big|
\Big\{
n\in{1,\ldots,N}:
|W_{n+s}-W_n|>\epsilon
\Big\}
\Big| 
=\,
\mathbb{P}_{n\sim\mathrm{Unif}({1,\ldots,N})}
\left(
|W_{n+s}-W_n|>\epsilon
\right)
\to 0.
\end{aligned}
\label{eq}
\end{equation}
Thus, the reasoning lengths for which
$|W_{n+s}-W_n|>\epsilon$
form an asymptotically zero-density set. This completes the proof.

\section{Proof of \Cref{general_prop}}
\label{ap:proof_general_pcp}

Recall from \Cref{eq:structured_gen_relative} that, for reasoning
sequences of lengths $n+s$ and $n$,
\begin{equation}
\log
\frac{
    \tilde P(y^*\mid x,Z_{n+s};\theta)
}{
    \tilde P(y^*\mid x,Z_n;\theta)
}
=
W_n-W_{n+s}
+
\rho_{n+s}(Z_{n+s})-\rho_n(Z_n).
\label{eq:general_pcp_decomposition}
\end{equation}

Let $L_N\sim\mathrm{Unif}(\{1,\ldots,N\})$. Substituting
$n=L_N$ into \Cref{eq:general_pcp_decomposition}, for any
$\epsilon>0$, the triangle inequality and union bound give
\begin{equation}
\begin{aligned}
&\mathbb{P}
\left(
\left|
\log
\frac{
    \tilde P(y^*\mid x,Z_{L_N+s};\theta)
}{
    \tilde P(y^*\mid x,Z_{L_N};\theta)
}
\right|
>\epsilon
\right) \\
&\quad\leq
\mathbb{P}
\left(
|W_{L_N}-W_{L_N+s}|>\frac{\epsilon}{2}
\right) 
+
\mathbb{P}
\left(
\left|
\rho_{L_N+s}(Z_{L_N+s})
-\rho_{L_N}(Z_{L_N})
\right|
>\frac{\epsilon}{2}
\right).
\end{aligned}
\label{eq:pcp_union_bound}
\end{equation}

We consider the two terms separately. By \Cref{entropy_stability_lemma},
the set of reasoning lengths satisfying
$|W_{n+s}-W_n|>\epsilon/2$ has asymptotically zero density. Since
$L_N$ is uniformly distributed over $\{1,\ldots,N\}$,
\begin{equation}
\begin{aligned}
&\mathbb{P}
\left(
|W_{L_N}-W_{L_N+s}|>\frac{\epsilon}{2}
\right) \\
&\quad=
\frac{1}{N}
\left|
\left\{
n\in\{1,\ldots,N\}:
|W_{n+s}-W_n|>\frac{\epsilon}{2}
\right\}
\right|
\to 0.
\end{aligned}
\label{eq:length_shift_prob}
\end{equation}

For the residual term, define
\begin{equation}
p_n(\epsilon)
:=
\mathbb{P}
\left(
\left|
\rho_{n+s}(Z_{n+s})-\rho_n(Z_n)
\right|
>\frac{\epsilon}{2}
\right).
\end{equation}
By assumption,
$p_n(\epsilon)\to0$ as $n\to\infty$. Averaging over the auxiliary
length $L_N$ gives
\begin{equation}
\mathbb{P}
\left(
\left|
\rho_{L_N+s}(Z_{L_N+s})
-\rho_{L_N}(Z_{L_N})
\right|
>\frac{\epsilon}{2}
\right) 
=
\frac{1}{N}\sum_{n=1}^{N}p_n(\epsilon)
\to0,
\label{eq:residual_cesaro}
\end{equation}
where the last step follows from Ces\`aro convergence.

Combining \Cref{eq:length_shift_prob,eq:residual_cesaro} with
\Cref{eq:pcp_union_bound}, we obtain
\begin{equation}
\lim_{N\to\infty}
\mathbb{P}
\left(
\left|
\log
\frac{
    \tilde P(y^*\mid x,Z_{L_N+s};\theta)
}{
    \tilde P(y^*\mid x,Z_{L_N};\theta)
}
\right|
>\epsilon
\right)
=0.
\end{equation}
This completes the proof.

\section{Diagnostic Experiment: Relation between Uncertainty and Concentration}\label{ap:uncertainty_concentration}
\begin{wrapfigure}{r}{0.5\linewidth}
    \centering
    \includegraphics[width=0.98\linewidth]{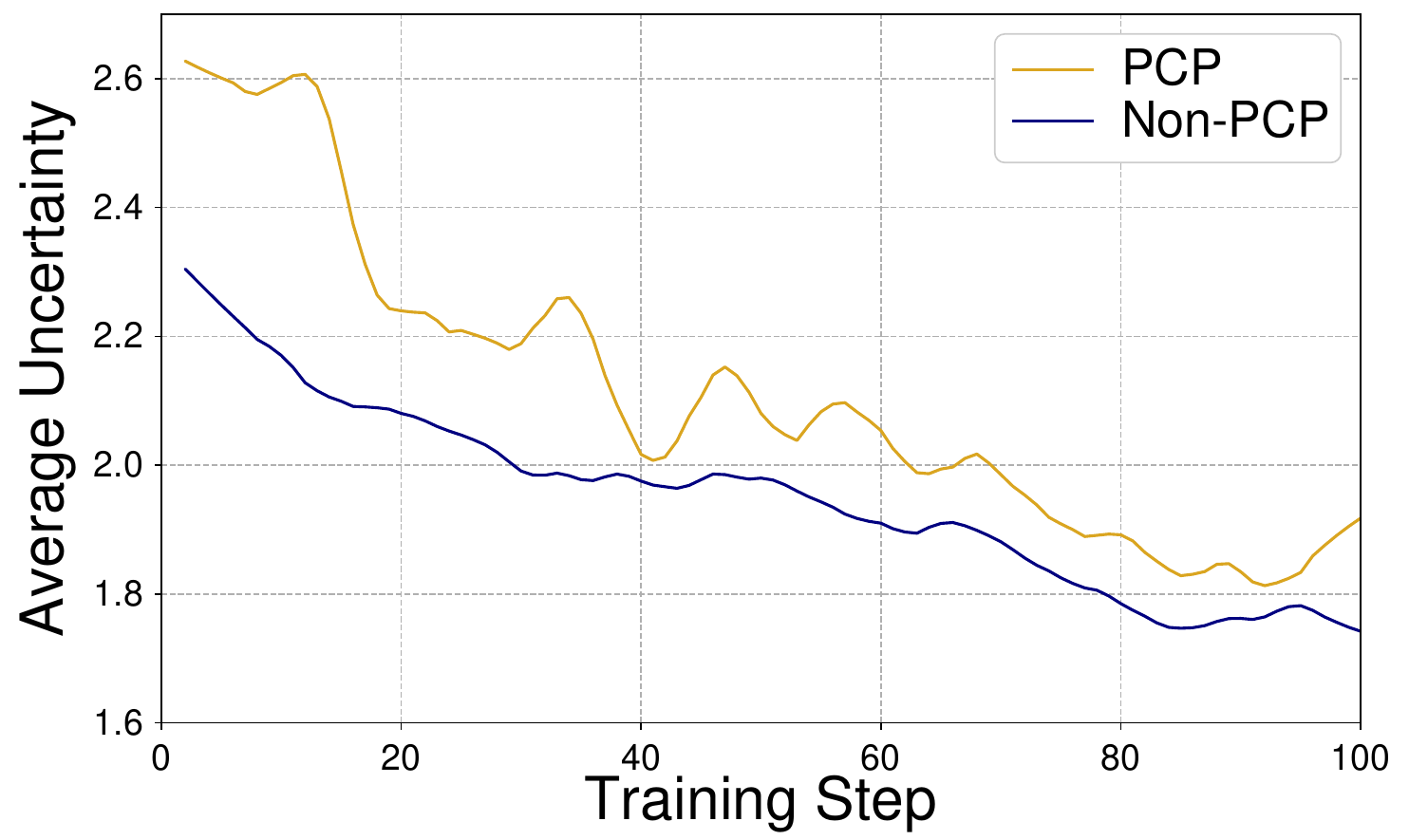}
    \caption{\textbf{Relation between uncertainty and concentration.} Throughout training, questions whose rollout groups satisfy the PCP condition %
    consistently exhibit higher average direct-answer uncertainty than non-PCP questions. This finding motivates uncertainty-aware data sampling.}
    \label{fig:exploration}
    \vspace{-0.12in}
\end{wrapfigure}
We investigate whether direct-answer uncertainty in \Cref{eq:uncertainty} can identify examples that are prone to posterior concentration. We divide the training examples into two groups according to the binary indicator of concentration in \Cref{eq:length_penalty} and compute the average direct-answer uncertainty in each group. \Cref{fig:exploration} shows that, along the training process, questions whose resulting rollout groups satisfy the PCP condition consistently exhibit higher average uncertainty than those that do not. Thus, pre-reasoning uncertainty is associated with subsequent reward concentration and can serve as a practical proxy for concentration risk. This finding motivates uncertainty-aware data sampling: preferentially retaining lower-uncertainty examples can reduce concentration-prone rollouts and preserve more informative posterior-reward differences. Appx.~\ref{ap:relation_exploration} provides the experimental details of \Cref{fig:exploration}, and Appx.~\ref{ap:additional_ablation} contains ablation study about the effect of different filtering rates.

\section{Complexity Analysis}\label{ap:complexity}
We compare the generation cost of uncertainty-aware data
sampling (UDS) with post-generation reward filtering.
Let $L=|x|$ denote the prompt length, $n$ the reasoning
length, and $k$ the number of tokens used to estimate
direct-answer uncertainty. We use an attention-cost
approximation with KV caching, suppressing factors
associated with model size. Generating $m$ tokens after
a prompt of length $L$ then costs
$\mathcal{O}(L^2 + Lm + m^2)$.

\paragraph{Uncertainty estimation.}
UDS estimates uncertainty using $k$ direct-answer tokens
and refreshes the cached estimates every $f$ training
steps. Amortizing each refresh over these steps gives
the per-question estimation cost
\begin{equation}
    C_{\mathrm{unc}}
    = \mathcal{O}\!\left(
        \frac{L^2 + Lk + k^2}{f}
      \right).
\end{equation}
This amortization assumes that the refresh cost is
distributed across the questions processed during
the corresponding interval.

\paragraph{Expected rollout cost.}
The expected fraction of questions retained by UDS is
\begin{equation}
    r = \sum_{i=1}^{3}
        (1-\omega_i)(\tau_i-\tau_{i-1}),
    \qquad \tau_0=0,\quad \tau_3=1.
\end{equation}
For a fixed rollout group size, the expected generation
cost per candidate question is therefore
\begin{equation}
    C_{\mathrm{UDS}}
    = \mathcal{O}\!\left(
        \frac{L^2 + Lk + k^2}{f}
        + r(L^2 + Ln + n^2)
      \right).
\end{equation}
In contrast, post-generation reward filtering generates
rollouts for every candidate question before deciding
which groups to discard, incurring
\begin{equation}
    C_{\mathrm{post}}
    = \mathcal{O}(L^2 + Ln + n^2).
\end{equation}

Thus, UDS reduces the rollout-generation term by the
retention factor $r$, at the expense of uncertainty
estimation. Under this approximation, net savings arise
when the amortized estimation overhead is smaller than
the avoided generation cost. Our default thresholds and
filtering rates give $r=0.55$, so approximately 45\% of
candidate questions are filtered before rollout generation.
This analysis concerns generation cost; it does not
quantify end-to-end training time, including reward
scoring and policy updates.

\section{Implementation Details}
\subsection{PCP's Exploration-Study Setup}\label{ap:exploration_setup}

We analyze posterior concentration using Qwen3-4B-Base on MMLU-Pro and GPQA-Diamond. For each dataset, we randomly sample 128 questions and generate eight responses per question, yielding 1,024 $(x,y,z)$ instances. We use a maximum response length of 1,536 tokens.

For \Cref{fig:mmlupro_posterior,fig:gpqa_posterior}, we partition the traces according to answer correctness, $\mathbbm{1}\{y=y^*\}$. We then fit separate third-order polynomial regressions of the posterior reward $P(y^*\Cond x,z)$ against reasoning length for the correct and incorrect traces and report 95\% confidence intervals.

For \Cref{fig:mmlupro_var}, we divide the 1,024 responses into 16 bins of width 100 tokens. The $k$-th bin contains traces satisfying $100(k-1)\leq |z|<100k$. We compute the posterior-reward variance within each bin and regress these values against the corresponding bin centers, $100k-50$.

For \Cref{fig:gpqa_var}, we preserve the rollout-group structure used by GRPO. For each rollout group $G$, we compute its average reasoning length and within-group posterior-reward variance as
\[
    \bar{\ell}_G=\frac{1}{|G|}\sum_{z\in G}|z|,
    \qquad
    v_G=\operatorname{Var}_{z\in G}
    \left[P(y_G^*\Cond x_G,z)\right].
\]
We then partition the 128 rollout groups into bins of width 100 tokens according to $\bar{\ell}_G$. For each non-empty bin, we plot the average $\bar{\ell}_G$ against the average $v_G$ over all groups in that bin and fit the displayed regression trend with a 95\% confidence interval. This analysis directly measures how posterior-reward dispersion within GRPO groups changes with their average reasoning length.
\subsection{Uncertainty--Concentration Study Setup}\label{ap:relation_exploration}
In \Cref{fig:exploration}, we examine the relationship between direct-answer uncertainty and posterior concentration using Qwen3-4B-Base. We disable the standard-deviation-based filtering used in RLPR so that concentration-prone rollout groups remain observable. At each training step, we partition the sampled questions into PCP and non-PCP subsets according to whether their resulting rollout groups activate the binary PCP indicator in \Cref{eq:length_penalty}. We then compute the average direct-answer uncertainty within each subset. For this diagnostic experiment, we train the model for 100 steps with a batch size of 320 and a GRPO group size of 8. Lastly, we report the subset-wise uncertainty averages throughout training. Additional ablations on the uncertainty-dependent filtering rates are provided in Appx.~\Cref{ap:additional_ablation}.

\subsection{RLCPR: Implementation Details}\label{ap:main_details}
This section provides additional implementation details for the main experiments in \Cref{subsec:main_results}. Similar to \cite{yu2025rlpr}, the policy model in RLCPR
generates 8 responses per question, using a learning rate of $1e-6$. The KL divergence
term is removed by setting the KL coefficient to 0. The batch size is set to 768 with 4 policy updates per step. The training prompt is provided in \Cref{ap:prompt}.

\subsection{RLCPR training prompt}\label{ap:prompt}
We provide the complete system prompt used in RLCPR training and validation:
\begin{promptbox}{RLCPR training prompt for Qwen3}
You are an assistant that solves problems by reasoning step by step.
You must always output your reasoning process inside <think> </think> tags,
and then your final answer inside <answer> </answer> tags.
Do not output anything else outside these tags.

Example:

<think>
The cost per jar of silver polish is $2.20. Fixed avoidable cost is $8,800.
Number of jars = 8,800 / 2.20 = 4,000.
</think>
<answer> 4,000 jars</answer>

Now follow the same format to answer the user's question.
user: {question}
Assistant:
\end{promptbox}
Specifically, \texttt{\{question\}} refers to the input question. To follow the requested output format, we remove the special tokens \texttt{<think>} and \texttt{</think>} from the Qwen3's vocabulary. 
\section{Additional Ablation Study}\label{ap:additional_ablation}
\paragraph{CR gate.}
Table~\ref{tab:ablation-gate} compares concentration-gated
regularization with applying the length penalty to all
groups. Removing the gate reduces accuracy by 1.6,
and 3.5 percentage points on MMLU-Pro and TheoremQA, respectively, while obtaining only comparable points on WebInstruct. These results support
selectively penalizing groups exhibiting posterior
concentration: applying the penalty indiscriminately
may discourage useful reasoning in groups whose
posterior rewards remain informative.

\noindent{\bf Uncertainty distribution.} \Cref{tab:ablation_distribution} demonstrates the ablation study about changing the hyper-parameters of uncertainty-aware data sampling (UDS). Specifically, in the first two rows, we reverse the order of the filter rates. These results show that filtering more high-uncertainty data provides substantial improvement over filtering more low-uncertainty data, echoing our diagnostic experiment in Appx.~\ref{ap:uncertainty_concentration}. In the last row, we increase the cutoff range, where fewer data are considered with particularly high or low uncertainty. In other words, most of the data are placed in the same bin, and the UDS does not essentially change the data distribution. This experiment also reduces the accuracy across all three mathematical benchmarks. It shows that reshaping the data distribution with direct-answer uncertainty effectively boosts the optimization performance.

\input{tab/ablation_II}
\noindent{\bf Regularization strength.} \Cref{tab:ablation_second} shows the ablation study on the strengths of the concentration-aware regularization. Specifically, we change the weight of the regularization reward, \emph{i.e.} $\lambda$. Overall, we recommend an optimal weight around $0.5$, while a particularly high or low value of $\lambda$ typically results in model degradation, possibly because of over or insufficient regularization. Notably, we fix the percentile levels to $(p_\delta,p_\alpha)=(0.8,0.2)$ throughout our experiments without hyperparameter tuning. The reported gains thus demonstrate the effectiveness of this default configuration, while model- or benchmark-specific tuning may further improve performance.

\section{Discussion about PCP's Residual Condition}\label{ap:discussion}
Our PCP analysis explicitly adopts convergence of the cross-trace residual difference as a sufficient condition. This assumption is motivated by \cite{bell2024slaves} and \cite{yang2026llmprobabilityconcentrationalignment}, who together establish convergence of the length-normalized difference between negative log-likelihood and realized entropy, alongside empirical evidence of diminishing deviations and variability as generation length increases. Connecting these findings to posterior rewards offers a useful perspective on verifier-free RL, although establishing when the required residual differences converge remains an open theoretical question beyond our scope. Our contribution is to characterize conditions under which posterior rewards become less discriminative across reasoning traces and examine their implications for policy optimization. Guided by this perspective, RLCPR accounts for concentration through uncertainty-aware sampling and selective regularization to improve training efficiency and performance. Future work may strengthen this foundation by deriving sufficient conditions and finite-length bounds for residual-difference convergence.

%% file: tab/ablation_II.tex
\begin{wraptable}{r}{0.48\textwidth}
    \vspace{-0.1in}
    \centering
    \captionsetup{font=small,skip=2pt}
    \footnotesize
    \setlength{\tabcolsep}{2pt}
    \renewcommand{\arraystretch}{1.15}

    \captionof{table}{Ablation of the CR gate.}
    \label{tab:ablation-gate}

    \resizebox{0.96\linewidth}{!}{%
    \begin{tabular}{l|ccc}
        \thicktoprule
        \textbf{Setting}
            & \textbf{MMLU-Pro}
            & \textbf{TheoremQA}
            & \textbf{WebInst.} \\
        \hline
        w/o CR gate & 59.7 & 57.6 & 79.0 \\
        Full model
            & \textbf{61.3}
            & \textbf{61.1}
            & \textbf{78.9} \\
        \thickbottomrule
    \end{tabular}%
    }

    \vspace{4pt}

    \captionof{table}{Ablation of UDS sampling settings.}
    \label{tab:ablation_distribution}

    \resizebox{\linewidth}{!}{%
    \begin{tabular}{cc|ccc}
        \thicktoprule
        \textbf{Cutoff}
            & \textbf{Filter rates}
            & \multirow{2}{*}{\textbf{MATH-500}}
            & \multirow{2}{*}{\textbf{Minerva}}
            & \multirow{2}{*}{\textbf{AIME24}} \\
        \(\boldsymbol{(\tau_1,\tau_2)}\)
            & \(\boldsymbol{(\omega_1,\omega_2,\omega_3)}\)
            & & & \\
        \hline
        \((0.3,0.7)\) & \((0.7,0.45,0.2)\)
            & 67.5 & 47.7 & 10.2 \\
        \((0.3,0.7)\) & \((0.2,0.45,0.7)\)
            & 82.9 & 64.5 & 16.3 \\
        \((0.1,0.9)\) & \((0.2,0.45,0.7)\)
            & 81.0 & 58.4 & 12.9 \\
        \thickbottomrule
    \end{tabular}%
    }

    \vspace{4pt}

    \captionof{table}{Ablation of CR strength.}
    \label{tab:ablation_second}

    \resizebox{0.92\linewidth}{!}{%
    \begin{tabular}{c|ccc}
        \thicktoprule
        \textbf{Reg. strength}
            & \textbf{MMLU-Pro}
            & \textbf{TheoremQA}
            & \textbf{WebInst.} \\
        \hline
        \(\lambda=0.2\) & 58.4 & 55.2 & 73.9 \\
        \(\lambda=0.5\) & 61.3 & 61.1 & 78.9 \\
        \(\lambda=1.0\) & 59.5 & 59.5 & 76.2 \\
        \thickbottomrule
    \end{tabular}%
    }

    \vspace{-0.1in}
\end{wraptable}